# Estimating the Hessian by Back-propagating Curvature

James Martens                                                                                         JMARTENS@CS.TORONTO.EDU
Ilya Sutskever                                                                                         ILYA@CS.UTORONTO.CA
Kevin Swersky                                                                                         KSWERSKY@CS.TORONTO.EDU

## Abstract

In this work we develop Curvature Propagation (CP), a general technique for efficiently computing unbiased approximations of the Hessian of any function that is computed using a computational graph. At the cost of roughly two gradient evaluations, CP can give a rank-1 approximation of the whole Hessian, and can be repeatedly applied to give increasingly precise unbiased estimates of any or all of the entries of the Hessian. Of particular interest is the diagonal of the Hessian, for which no general approach is known to exist that is both efficient and accurate. We show in experiments that CP turns out to work well in practice, giving very accurate estimates of the Hessian of neural networks, for example, with a relatively small amount of work. We also apply CP to Score Matching, where a diagonal of a Hessian plays an integral role in the Score Matching objective, and where it is usually computed exactly using inefficient algorithms which do not scale to larger and more complex models.

## 1. Introduction

There are many models and learning algorithms where it becomes necessary, or is at least very useful, to compute entries of the Hessian of some complicated function. For functions that can be computed using a computational graph there are automatic methods available for computing Hessian-vector products exactly (e.g. Pearlmutter, 1994). These can be used to recover specific columns of the Hessian, but are inefficient at recovering other parts of the matrix such as large blocks, or the diagonal. For the diagonal of the Hessian of a neural network training objective, there are deterministic approximations available such as that of Becker and Le Cun (1988), but these are not guaranteed to be accurate.

Recently Chapelle and Erhan (2011) showed how to compute an unbiased estimate of the diagonal of the Gauss-Newton matrix, and used this to perform preconditioning within a Hessian-free Newton optimization algorithm (Martens, 2010). In this paper we build upon this idea and develop a family of algorithms, which we call Curvature Propagation (CP), for efficiently computing unbiased estimators of the Hessians of arbitrary functions. Estimating entries of the Hessian turns out to be strictly harder than doing the same for the Gauss-Newton matrix, and the resulting approach is necessarily more complex, requiring several additional ideas.

As with the algorithm of Chapelle and Erhan (2011), CP involves reverse sweeps of the computational graph of the function, which can be repeated to obtain higher-rank estimates of arbitrary accuracy. And when applied to a function which decomposes as the sum of $M$ terms, such as typical training objective functions, applying CP to the terms individually results in an estimate of rank $M$, at no additional expense than applying it to the sum.

This is useful in many applications. For example, the diagonal of the Hessian can be used as a preconditioner for first and second order nonlinear optimizers, which is the motivating application of Becker and Le Cun (1988) and Chapelle and Erhan (2011). Another example is Score Matching (Hyvarinen, 2006), a method for parameter estimation in Markov Random Fields. Because Score Matching uses the diagonal of the Hessian within its objective it is expensive to apply the method to all but the simplest models. As we will see, CP makes it possible to efficiently apply Score Matching to any model.

## 2. Derivation of CP

In the following section we develop the Curvature Propagation method (CP) for functions that are defined in terms of general computational graphs. We will present one version of the approach that relies on the use of complex arithmetic, and later also give a version that uses only real arithmetic.

At a high level, we will define complex vector-valued linear function on the computational graph of our target function $f$, and then show through a series of lemmas that the expectation of the self outer-product of this function is in fact the Hessian matrix. This function can be computed by what amounts to a modification of reverse-mode automatic differentiation, where noise is injected at each node.

## 2.1. Setting and notation

Let $f : \mathbb{R}^n \longrightarrow \mathbb{R}$ be a twice differentiable function. We will assume that $f$ can be computed via a computation graph consisting of a set of nodes $N = \{i : 1 \leq i \leq L\}$ and directed edges $E = (i,j) : i, j \in N$, where at each node $i$ there is a vector valued output $y_i \in \mathbb{R}^{n_i}$ is computed via $y_i = f_i(x_i)$ for some twice-differentiable function $f_i$. Here $x_i \in \mathbb{R}^{m_i}$ is the total input to node $i$, and is given by the concatenation of vectors $y_k$ for $k \in P_i$ and $P_i = \{k : k \text{ is a parent of } i\} = \{k : (k,i) \in E\}$. We identify node 1 as input or "source" node (so that $P_1 = \emptyset$) and node $L$ as the output or "sink" node, with $y_L = f(y_1)$ being the final output of the graph.

Let $J_b^a$ denote the Jacobian of $a$ w.r.t. $b$ where $a$ and $b$ are vectors, or in other words, $\frac{\partial a}{\partial b}$. And let $H_{a,b}^c$ denote the Hessian of the scalar function $c$ w.r.t. $a$ and then w.r.t. $b$ (the order matters since it determines the dimension of the matrix). Note that if $a$ and $b$ are quantities associated with nodes $i$ and $j$ (resp.) in the computational graph, $J_b^a$ and $H_{a,b}^c$ will only be well-defined when $j$ does not depend directly or indirectly on $i$, i.e. $i \notin A_j$, where $A_j = \{k : k \text{ is an ancestor of } j\}$. Also note that when there is no dependency on $b$ of $a$ it will be the case that $J_b^a = 0$. Under this notation, the Hessian of $f$ w.r.t. its input is denoted by $H_{y_1,y_1}^f$, but we will use the short-hand $H$ for convenience.

For $k \in P_i$, let $R_{i,k}$ denote the projection matrix which maps the output $y_k$ of node $k$ to the their positions in node $i$'s input vector $x_i$, so that we have $x_i = \sum_{k \in P_i} R_{i,k} y_k$.

Summarizing, we have the following set of recursive definitions for computing $y_L = f(y_1)$ which are iterated for $i$ ranging from 2 to $L$:

$$x_i = \sum_{k \in P_i} R_{i,k} y_k \qquad y_i = f_i(x_i)$$

Note that $R_{i,k}$ need not appear explicitly as a matrix when implementing these recursions in actual code, but is merely the formal mathematical representation we will use to describe the projective mapping which is performed whenever outputs from a given computational node $k$ are used as input to another node $i$.

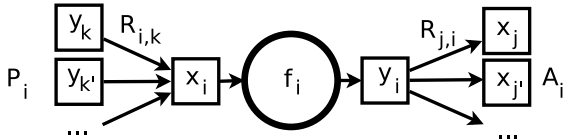

## 2.2. Computing gradients and Hessians

Reverse-mode automatic differentiation[1] is a well known method for computing the gradient of functions which are

[1] also known as back-propagation (Rumelhart et al., 1986) in the context of neural networks

defined in terms of computation graphs. It works by starting at the final node $L$ and going backwards through the graph, recursively computing the gradient of $f$ w.r.t. the $y_i$ for each $i$ once the same has been done for all of $i$'s children. Using the vector-valued computational graph formalism and notation we have established, the recursions for computing the gradient $\nabla f = J_{y_1}^f$ (remembering that $f \equiv y_L$) are given by

$$J_{y_L}^f = 1 \tag{1}$$

$$J_{y_i}^f = \sum_{k \in C_i} J_{x_k}^f J_{y_i}^{x_k} = \sum_{k \in C_i} J_{x_k}^f R_{k,i}^\top \tag{2}$$

$$J_{x_i}^f = J_{y_i}^f J_{x_i}^{y_i} \tag{3}$$

where $C_i = \{k : k \text{ is a child of } i\}$ and we have used the fact that $J_{y_i}^{x_k} = R_{k,i}^\top$.

For this method to yield a realizable algorithm, it is assumed that for each node $i$, the function $f_i$ is simple enough that direct computation of and/or multiplication by the "local" Jacobian $J_{x_i}^{y_i} = f_i'(x_i)$ is easy. If for a particular node $i$ this is not the case, then the usual procedure is to split $i$ into several new nodes which effectively break $f_i$ into several computationally simpler pieces.

By computing the vector derivative of both sides of each of the above equations w.r.t. $y_L$, $y_j$, and $y_j$ respectively (for $j \notin A_i$), the following recursions can be derived

$$H_{y_L,y_L}^f = 0 \tag{4}$$

$$H_{y_i,y_j}^f = \sum_{k \in C_i} R_{k,i} H_{x_k,y_j}^f \tag{5}$$

$$H_{x_i,y_j}^f = J_{x_i}^{y_i\top} H_{y_i,y_j}^f + M_i J_{y_j}^{x_i} \tag{6}$$

where

$$M_i \equiv \sum_{q=1}^{n_i} J_{y_{i,q}}^f H_{x_i,x_i}^{y_{i,q}} \tag{7}$$

and where $y_{i,q}$ denotes the $q$-th component of $y_i$. In deriving the above it is important to remember that $R_{k,i}$ is a constant, so that its Jacobian w.r.t. $y_j$ is the zero matrix. Also note that $J_{y_{i,q}}^f$ is a scalar and that $H_{x_i,x_i}^{y_{i,q}}$ is the Hessian of the local nonlinearity $f_i$. The overall Hessian of $f$, $H_{y_1,y_1}^f$ can be obtained by applying these recursions in a backwards manner (assuming that the various Jacobians are already computed).

The additional Jacobian terms of the form $J_{y_j}^{x_i}$ which appear in eqn. 6 can be computed according to recursions analogous to those used to compute the gradient, which are given by the equations below:

$$J_{x_i}^{x_i} = I_{m_i \times m_i} \tag{8}$$

$$J_{x_j}^{x_i} = J_{y_j}^{x_i} J_{x_j}^{y_j} \tag{9}$$

$$J_{y_j}^{x_i} = \sum_{k \in C_j} J_{x_k}^{x_i} J_{y_j}^{x_k} = \sum_{k \in C_j} J_{x_k}^{x_i} R_{k,j} \quad \forall i \notin A_j \tag{10}$$

$$J_{x_j}^{x_i} = 0 \quad \forall i \in A_j \tag{11}$$

where, for convenience, we have defined $J_{x_i, x_j}$ to be zero whenever $i$ is an ancestor of $j$, whereas otherwise it would be undefined.

In general, using these recursions for direct computation of $H^f_{y_1,y_1}$ will be highly impractical unless the computation tree for $f$ involves a small total number of nodes, each with small associated output and input dimensions $n_i$ and $m_i$. The purpose in giving them is to reveal how the "structure" of the Hessian follows the computation tree, which will become critically important in both motivating the CP algorithm and then proving its correctness.

### 2.3. The $S$ function

We now define an efficiently computable function $S$ that will allow us to obtain rank-1 estimates of the Hessian. Its argument consists of an ordered list of vectors $V \equiv \{v_i\}_{i=1}^L$ where $v_i \in \mathbb{R}^{\ell_i}$, and its output is a $n$-dimensional vector (which may be complex valued). It will be defined as $S(V) \equiv S_{y_1}(V)$, where $S_{y_i}(V) \in \mathbb{C}^{n_i}$ and $S_{x_i}(V) \in \mathbb{C}^{m_i}$ are vector-valued functions of $V$ defined recursively via the equations

$$S_{y_L}(V) = 0 \tag{12}$$

$$S_{y_i}(V) = \sum_{k \in C_i} R_{k,i}^\top S_{x_k}(V) \tag{13}$$

$$S_{x_i}(V) = F_i^\top v_i + J_{x_i}^{y_i\,\top} S_{y_i}(V) \tag{14}$$

where each $F_i$ is a (not necessarily square) complex-valued matrix in $\mathbb{C}^{\ell_i \times m_i}$ satisfying $F_i^\top F_i = M_i$. Such an $F_i$ is guaranteed to exist because $M_i$ is symmetric, which follows from the fact that it is a linear combination of Hessian matrices.

Note that these recursions closely resemble those given previously for computing the gradient (eqn. 1, 2, and 3). The multiplication by $J_{x_i}^{y_i\,\top}$ of the vector $S_{y_i}(V)$ at each stage of the recursion is easy to perform since this is precisely what happens at each stage of reverse-mode automatic differentiation used to compute the gradient of $f$. In general, the cost of computing $S$ is similar to that of computing the gradient, which itself is similar to that of evaluating $f$. The practical aspects computing $S(V)$ will be discussed further in section 5.

### 2.4. Properties of the $S$ function with stochastic inputs

Suppose that the random variable $V$ satisfies:

$$\forall i \ \mathbb{E}\left[v_i v_i^\top\right] = I \quad \text{and} \quad \forall j \neq i, \ \mathbb{E}\left[v_i v_j^\top\right] = 0 \tag{15}$$

For example, each $v_i$ could be drawn from a multivariate normal with mean 0 and covariance matrix $I$.

We will now give a result which establishes the usefulness of $S(V)$ as a tool for approximating $H$. The proof of this theorem and others will be located in the appendix/supplement.

**Theorem 2.1.** $\mathbb{E}[S(V)S(V)^\top] = H^f_{y_1,y_1}(\equiv H)$

In addition to being an unbiased estimator of $H$, $S(V)S(V)^\top$ will be symmetric and possibly complex-valued. To achieve a real-valued estimate we can instead use only the real component of $S(V)S(V)^\top$, which itself will also be an unbiased estimator for $H$ since the imaginary part of $S(V)S(V)^\top$ is zero in expectation.

### 2.5. Avoiding complex numbers

The factorization of the $M_i$'s and resulting complex arithmetic associated with using these factors can be avoided if we redefine $V$ so that each $v_i$ is of dimension $m_i$ (instead of $\ell_i$), and we define the real vector-valued functions $T(V) \equiv T_{y_1}(V)$ and $U(V) \equiv U_{y_1}(V)$ according to the following recursions:

$$T_{y_L}(V) = 0 \qquad\qquad U_{y_L}(V) = 0$$

$$T_{y_i}(V) = \sum_{k \in C_i} J_{y_i}^{x_k\,\top} T_{x_k}(V) \qquad U_{y_i}(V) = \sum_{k \in C_i} J_{y_i}^{x_k\,\top} U_{x_k}(V)$$

$$T_{x_i}(V) = M_i v_i + J_{x_i}^{y_i\,\top} T_{y_i}(V) \quad U_{x_i}(V) = v_i + J_{x_i}^{y_i\,\top} U_{y_i}(V)$$

Both these recursions for $T$ and $U$ are trivial modifications of those given for $S(V)$, with the only difference being the matrix which multiplies $v_i$ (it's $F_i^\top$ for $S$, $M_i$ for $T$, and $I$ for $U$). And because they do not involve complex quantities at any point, they will be real-valued.

**Theorem 2.2.** $T(V)U(V)^\top$ is an unbiased estimator of $H$

Since $H$ is symmetric, it follows directly from this result that $\left(T(V)U(V)^\top\right)^\top = U(V)T(V)^\top$ is also an unbiased estimator of $H$. Note however that while both $T(V)U(V)^\top$ and $U(V)T(V)^\top$ will be symmetric in expectation (since $H^f_{y_1,y_1}$ is), for any particular choice of $V$ they generally will not be. This issue can be addressed by instead using the estimator $\frac{1}{2}\left(T(V)U(V)^\top + U(V)T(V)^\top\right)$ which will be symmetric for any $V$. However, despite the fact that $S(V)S(V)^\top$ and this alternative estimator are both symmetric for all $V$'s and also unbiased, they will not, in general, be equal. While computing both $T$ and $U$ will require a total of 2 sweeps over the computational graph versus only the one required for $S(V)$, the total amount of work will be the same due to the doubly expensive complex-valued arithmetic required to evaluate $S(V)$.

### 2.6. Matrix interpretation of $S$, $T$ and $U$

Suppose we represent $V$ as a large vector $v \equiv [v_1^\top \ldots v_L^\top]^\top$ with dimension $m \equiv \sum_i m_i$. Then the functions $S$, $T$ and $U$ are linear in the $v_i$'s (a fact which follows from the recursive definitions of these functions) and hence $v$. Thus $S$, $T$, and $U$ have an associated representation as matrices $\tilde{S} \in \mathbb{C}^{n \times m}$, $\tilde{T} \in \mathbb{R}^{n \times m}$, and $\tilde{U} \in \mathbb{R}^{n \times m}$ w.r.t. the coordinate bases given by $\tilde{v}$.

Then noting that $S(V)S(V)^\top = \tilde{S}vv^\top \tilde{S}^\top$, and that condition (15) is equivalent to $\mathbb{E}[vv^\top] = I$, we obtain

$$H^f_{y_1,y_1} = \mathbb{E}\left[\tilde{S}vv^\top \tilde{S}^\top\right] = \tilde{S}\,\mathbb{E}\left[vv^\top\right]\tilde{S}^\top = \tilde{S}\tilde{S}^\top$$

and thus we can see that $\tilde{S}$ has an interpretation as a "factor" of $H^f_{y_1,y_1}$. Similarly we have $\tilde{T}\tilde{U}^\top = H^f_{y_1,y_1}$ and $\tilde{U}\tilde{T}^\top = H^f_{y_1,y_1}$.

## 3. A simpler method?

At the cost of roughly two passes through the computational graph it is possible to compute the Hessian-vector $Hw$ for an arbitrary vector $w \in \mathbb{R}^n$ (e.g. Pearlmutter, 1994). This suggests the following simple approach to computing an unbiased rank-1 estimate of $H$: draw $w$ from a distribution satisfying $E[ww^\top] = I$ and then take the outer product of $Hw$ with $w$. It is easy to see that this is unbiased, since

$$\mathbb{E}\left[Hww^T\right] = H\,\mathbb{E}\left[ww^T\right] = H \quad (16)$$

Computationally, this estimator is just as expensive as CP, but since there are several pre-existing methods computing Hessian vector products, it may be easier to implement. However, we will prove in the next section that the CP estimator will have much lower variance in most situations, and later confirm these findings experimentally. And in addition to this, there are certain situations, which arise frequently in machine learning applications, where vectorized implementations of CP will consume far less memory than similar vectorized implementations of this simpler estimator ever could, and we will demonstrate this in the specific case when $f$ is a neural network training objective function.

It is also worth noting that this estimator underlies the Hessian norm estimation technique used in Rifai et al. (2011). That this is true is due to the equivalence between the stochastic finite-difference formulation used in that work and matrix-vector products with randomly drawn vectors. We will make this rigorous in the appendix/supplement.

## 4. Covariance analysis

Let $AB^\top$ be an arbitrary matrix factorization of $H$, with $A, B \in \mathbb{C}^{n \times \ell}$. Given a vector valued random variable $u \in \mathbb{R}^\ell$ satisfying $E[uu^\top] = I$, we can use this factorization to produce an unbiased rank-1 estimate of the Hessian, $H^{A,B} \equiv (Au)(Bu)^\top = Auu^\top B^\top$. Note that the various CP estimators, as well as the simpler one discussed in the previous section are all of this form, and differ only in their choices of $A$ and $B$.

Expanding we have:

$$\mathbb{E}[H^{A,B}_{ij}H^{A,B}_{kl}] = \mathbb{E}\left[\sum_{a,b}A_{i,a}u_au_bB_{j,b}\sum_{c,d}A_{k,c}u_cu_dB_{l,d}\right] \quad (17)$$

$$= \sum_{a,b,c,d} A_{ia}B_{jb}A_{kc}B_{ld}\,\mathbb{E}\left[u_au_bu_cu_d\right] \quad (18)$$

where here (and in the remainder of this section) the subscripts on $u$ refer to scalar components of the vector $u$ and not elements of a collection of vectors.

If we assume $u \sim G \equiv \text{Normal}(0, I)$, we can use the well-know formula $\mathbb{E}_G[u_au_bu_cu_d] = \delta_{ab}\delta_{cd} + \delta_{ac}\delta_{bd} + \delta_{ad}\delta_{bc}$ and simplify this further to:

$$= \sum_{a,b,c,d} A_{ia}B_{jb}A_{kc}B_{ld}(\delta_{ab}\delta_{cd} + \delta_{ac}\delta_{bd} + \delta_{ad}\delta_{bc})$$

$$= (A_i^\top B_j)(A_k^\top B_l) + (A_i^\top A_k)(B_j^\top B_l) + (A_i^\top B_l)(A_k^\top B_j)$$

$$= H_{ij}H_{kl} + (A_i^\top A_k)(B_j^\top B_l) + H_{il}H_{jk}$$

where $A_i$ is a vector consisting of the $i$-th row of $A$, and similarly for $B_i$, and where we have used $H_{ij} = A_i^\top B_j$. Consequently, the variance is given by:

$$\text{Cov}_G\left[H^{A,B}_{ij}, H^{A,B}_{kl}\right] = \mathbb{E}_G\left[H^{A,B}_{ij}H^{A,B}_{kl}\right] - H_{ij}H_{kl}$$
$$= (A_i^\top A_k)(B_j^\top B_l) + H_{il}H_{jk}$$

Note that when $A = B = \tilde{S}$, we have that $(A_i^\top A_k)(B_j^\top B_l) = (\tilde{S}_i^\top \tilde{S}_k)(\tilde{S}_j^\top \tilde{S}_l) = H_{ik}H_{jl}$. Thus the estimator $H^{\tilde{S},\tilde{S}}$ has the following desirable property: its covariance depends only on $H$ and not on the specific details of the computational graph used to construct the $S$ function.

If on the other hand we assume that $u \sim K \equiv \text{Bernoilli}(\{-1,1\})^\ell$, i.e. $K$ is a multivariate distribution of independent Bernoulli random variables on $\{-1, 1\}$, we have $\mathbb{E}_B[u_au_bu_cu_d] = \delta_{ab}\delta_{cd} + \delta_{ac}\delta_{bd} + \delta_{ad}\delta_{bc} - 2\delta_{ab}\delta_{bc}\delta_{cd}$, which when plugged into (18) gives:

$$\text{Cov}_K\left[H^{A,B}_{ij}H^{A,B}_{kl}\right] = (A_i^\top A_k)(B_j^\top B_l) + H_{il}H_{jk}$$
$$- 2\sum_a B_{ia}A_{ja}B_{ka}A_{la}$$
$$= \text{Cov}_G\left[H^{A,B}_{ij}H^{A,B}_{kl}\right] - 2\sum_a B_{ia}A_{ja}B_{ka}A_{la}$$

Of particular interest is the self-variance of $H^{A,B}_{ij}$ (i.e. $\text{Var}\left[H^{A,B}_{ij}\right] = \text{Cov}\left[H^{A,B}_{ij}, H^{A,B}_{ij}\right]$). In this case we have that:

$$\text{Var}_K\left[H^{A,B}_{ij}\right] = \text{Var}_G\left[H^{A,B}_{ij}\right] - 2\sum_a (B_{ia}A_{ja})^2$$

and we see that variance of estimator that uses $K$ will always be strictly smaller than the one that uses $G$, unless $\sum_a (B_{ia}A_{ja})^2 = 0$ (which would imply that $\sum_a B_{ia}A_{ja} = H_{ij} = 0$).

Returning to the case that $u \sim G$, we can prove the following result, which shows that when it comes to estimating the diagonal entries $H_{ii}$ of $H$, the estimator which uses $A = B = \tilde{S}$ has the lowest variance among all possible estimators of the form $H^{A,B}$:

**Theorem 4.1.** $\forall i$ and $\forall A, B$ s.t. $AB^\top = H$ we have:

$$\text{Var}_G\left[H^{A,B}_{ii}\right] \geq \text{Var}_G\left[H^{\tilde{S},\tilde{S}}_{ii}\right] = 2H^2_{ii}$$

Moreover, in the particular case of using the 'simple' estimator (which is given by $A = H, B = I$) the variance of the diagonal entries is given by:

$$\text{Var}_G\left[H_{ii}^{H,I}\right] = H_i^\top H_i + H_{ii}^2 = \sum_{j \neq i} H_{ij}^2 + \text{Var}_G\left[H_{ii}^{\tilde{S},\tilde{S}}\right]$$

and so we can see that the CP estimator based on $S$ always gives a lower variance, and is strictly lower in most cases.

## 5. Practical aspects

### 5.1. Computing and factoring the $M_i$'s

Computing the matrices $M_i$ for each node $i$ is necessary in order to compute the $S$, $T$ and $U$ functions, and for $S$ we must also be able to factor them. Fortunately, each $M_i$ can be computed straightforwardly according to eqn. 7 as long as the operations performed at node $i$ are simple enough. And each $H_{x_i,x_i}^{y_i,q}$ is determined completely by the local function $f_i$ computed at node $i$. The Jacobian term $J_{y_{i,q}}^f = \left[J_{y_i}^f\right]_q$ which appears in the formula for $M_i$ is just a scalar, and is the derivative of $f$ w.r.t. $y_{i,q}$. This can be made cheaply and easily available by performing, in parallel with the computation of $S(V)$, the standard backwards automatic differentiation pass for computing the gradient of $f$ w.r.t. to $y_1$, which will produce the gradient of $f$ w.r.t. each $y_i$ along the way. Alternatively, this gradient information may be cached from a gradient computation which is performed ahead of time (which in many applications is done anyway).

In general, when $M_i$ is block or banded diagonal, $F_i$ will be as well (with the same pattern), which will greatly reduce the associated computational and storage requirements. For example, when $f_i$ corresponds to the element-wise nonlinearities computed in a particular layer of a neural network, $M_i$ will be diagonal and hence so will $F_i$, and these matrices can be stored as such. Also, if $M_i$ happens to be sparse or low rank, without any other obvious special structure, there are algorithms which can compute factors $F_i$ which will also be sparse or low-rank.

Alternatively, in the most extreme case, the vector valued nodes in the graph can be sub-divided to produce a graph with the property that every node outputs only a scalar and has at most 2 inputs. In such a case, each $M_i$ will be no bigger than $2 \times 2$. Such an approach is best avoided unless deemed necessary since the vector formalism allows for a much more vectorized and thus efficient implementation in most situations which arise in practice. Another option to consider if it turns out that $M_i$ is easy to work with but hard to factor, is to use the $T, U$ based estimator instead of the $S$ based one.

It may also be the case that $M_i$ or $F_i$ will have a special sparse form which makes sampling the entire vector $v_i$ unnecessary. For example, if a node copies a large input vector to its output and transforms a single entry by some non-linear function, $M_i$ will be all zeros with a single element on the diagonal (and hence so will its factor $F_i$), making it possible to sample only the component of $v_i$ that corresponds to that entry.

### 5.2. Increasing the rank

As with any unbiased estimator, the estimate can be made more accurate by collecting multiple samples. Fortunately, sampling and computing $S(V)$ for multiple $V$'s is trivially parallelizeable. And it can be easily implemented in vectorized code for $k$ samples by taking the defining recursions for $S$ (eqn. 12, 13, and 14) and redefining $S_{y_i}(V)$ and $S_{x_i}(V)$ to be matrix valued functions (with $k$ columns) and $v_i$ to be a $m_i \times k$ matrix of column vectors which are generated from independent draws from the usual distribution for $v_i$.

In the case where $f$ is a sum of $B$ similarly structured terms, which occurs frequently in machine learning such as when $f$ is sum of regression errors or log-likelihood terms over a collection of training cases, one can apply CP individually to each term in the sum at almost no extra cost as just applying it to $f$, thus obtaining a rank-k estimate of $f$ instead of a rank-1 estimate.

### 5.3. Curvature Propagation for Diagonal Hessian Estimation in Feedforward Neural Networks

In this section we will apply CP to the specific example of computing an unbiased estimate $\text{diag}(\hat{H})$ of the diagonal of $H$ ($\text{diag}(H)$) of a feed-forward neural networks with $\ell$ layers. The pseudocode below computes the objective $f$ of our neural network for a batch of $B$ cases.

1: Input: $z_1$, a matrix of inputs (with $B$ columns, one per case)
2: **for all** $i$ **from** 1 to $\ell - 1$ **do**
3:     $u_{i+1} \leftarrow W_i z_i$
4:     $z_{i+1} \leftarrow g(u_{i+1})$
5: **end for**
6: $f \leftarrow \sum_{b=1}^{B} L_b(z_{\ell,b})/B$
7: Output: $f$

Here $g(x)$ is a coordinate-wise nonlinearity, $z_i$ are matrices containing the outputs of the neuronal units at layer $i$ for all the cases, and similarly the matrices $u_i$ contain their inputs. $L_b$ denotes the loss-function associated with case $b$ (the dependency on $b$ is necessary so we can include targets). For simplicity we will assume that $L_b$ is the standard squared loss given by $L_b(z_{\ell,b}) = 1/2\|z_{\ell,b} - t_b\|^2$ for target vector $t_b$ (where $t$ will denote the matrix of these vectors).

The special structure of this objective permits us to efficiently apply CP to each scalar term of the average $\sum_{b=1}^{B} L_b(z_{\ell,b})/B$, instead of to $f$ directly. By summing the estimates of the diagonal Hessians for each $L_b(z_{\ell,b})/B$ we thus obtain a rank-$B$ estimate of $H$ instead of merely a rank-1 estimate. That this is just as efficient as applying CP directly to $f$ is due to the fact that the computations of each $z_{\ell,b}$ are performed independently of each other.

For ease of presentation, we will redefine $V \equiv \{v_i\}_{i=1}^{L}$ so that each $v_i$ is not a single vector, but a matrix of such vectors with $B$ columns. We construct the computational graph so that the element-wise nonlinearities and the weight matrix multiplications performed at each of the $\ell$ layers each correspond to a node in the graph. We define $S_{u_i} \equiv S_{x_{j_i}}(V)$ where $j_i$ is the node corresponding to the computation of $u_i$ (from $z_{i-1}$ and $W_{i-1}$), $S_{z_i} \equiv S_{x_{k_i}}(V)$ where $k_i$ is the node correspond to the computation of $z_i$ (from $u_i$), and $S_{W_i} \equiv [S_{y_1}(V)]_{W_i}$ where $[\cdot]_{W_i}$ denotes extraction of the rows in $y_1$ corresponding to the $i$-th weight-matrix ($W_i$). The variables $d_{z_i}$ and $d_{u_i}$ are derivatives w.r.t. $u_i$ and $z_i$ and are computed with backpropagation. Consistent with our mild redefinition / abuse of notation for $V$, each of $S_{u_i}$, $S_{z_i}$ and $S_{W_i}$ will be matrix-valued with a column for each of the $B$ training cases. Finally, we let $a \odot b$ denote the element-wise product, $a^{\odot 2}$ the element-wise power, $\text{outer}(a, b) \equiv ab^\top$, $\text{outer2}(a, b) \equiv \text{outer}(a^{\odot 2}, b^{\odot 2})$, and $\text{vec}(\cdot)$ the vectorization operator.

Under this notation, the algorithm below estimates the diagonal of the Hessian of $f$ by estimating the sub-objective corresponding to each case, and then averaging the results. Like the pseudo-code for the neural network objective itself, it makes use of vectorization, which allows for an easily parallelized implementation.

1: $S_{z_\ell} \leftarrow v_{j_\ell}$ ; $d_{z_\ell} \leftarrow z_\ell - t$
2: $S_{u_\ell} \leftarrow S_{z_\ell}$ ; $d_{u_\ell} \leftarrow d_{z_\ell}$
3: **for all** $i$ **from** $\ell - 1$ **down to** $1$ **do**
4: $\quad S_{z_i} \leftarrow W_i^\top S_{u_{i+1}}$ ; $d_{z_i} \leftarrow W_i^\top d_{u_{i+1}}$
5: $\quad [\text{diag}(\hat{H})]_{W_i} \leftarrow \text{vec}(\text{outer2}(z_i, S_{u_{i+1}}))/B$
6: $\quad K_i \leftarrow g''(u_i) \odot d_{z_i}$
7: $\quad S_{u_i} \leftarrow S_{z_i} \odot g'(u_i) + v_{k_i} \odot K_i^{\odot 1/2}$
8: $\quad d_{u_i} \leftarrow d_{z_i} \odot g'(u_i)$
9: **end for**

For $i < \ell$, each $K_i$ is a $B$-columned matrix of vectors containing the diagonals for each training case of the local matrices $M_{k_i}$ for each case occurring at node $k_i$. Because $M_j$ corresponds to an element-wise non-linear function, it is diagonal, and so $K_i^{\odot 1/2}$ will be a matrix of vectors corresponding to the diagonals of the factors $F_{k_i}$ (which are themselves diagonal). Note that the above algorithm makes use of the fact that the local matrices $M_{j_i}$ can be set to zero and the estimator of the diagonal will remain unbiased.

At no point in the above implementation do we need to store any matrix the size of $S_{W_i}$, as the computation of $[\text{diag}(\hat{H})]_{W_i}$, which involves an element-wise square of $S_{W_i}$ and a sum over cases (as accomplished by line 5), can be performed as soon as the one and only contribution to $S_{W_i}$ from other nodes in the graph is available. This is desirable since $S_{W_i}$ will usually be much larger than the various other intermediate quantities which we need to store, such as $z_i$ or $S_{u_i}$. In functions $f$ where large groups of parameters are accessed repeatedly throughout the computation graph, such as in the training objective of recurrent neural networks, we may have to temporally store some matrices the size $S_{y_1}$ (or certain row-restrictions of this, like $S_{W_i}$) as the contributions from different cases are collected and summed together, which can make CP less practical. Notably, despite the structural similarities of backprop (BP) to CP, this problem doesn't exist with BP since one can store incomplete contributions from each case in the batch into a single $n$ dimensional vector, which is impossible in CP due to the need to take the entry-wise square of $S_{y_1}$ before summing over cases.

## 6. Hardness of exact computation

An approach like CP wouldn't be as useful if there was an efficient and exact algorithm for computing the diagonal of the Hessian of the function defined by an arbitrary computation graph. In this section we will argue why such an algorithm is unlikely to exist.

To do this we will reduce the problem of multiplying two matrices to that of computing (exactly) the diagonal of the Hessian of a certain function $f$, and then appeal to a hardness due to Raz and Shpilka (2001) which shows that matrix multiplication will require asymptotically more computation than CP does when it is applied to $f$. This result assumes a limited computational model consisting of bounded depth arithmetic circuits with arbitrary fan-in gates. While not a fully general model of efficient computation, it nonetheless captures most natural algebraic formulae and algorithms that one might try to use to compute the diagonal of $f$.

The function $f$ will be defined by: $f(y) \equiv 1/2 y^\top W^\top Z W y$, where $Z \in \mathbb{R}^{2n \times 2n}$ is symmetric, and $W \equiv [P^\top Q]^\top$ with $P \in \mathbb{R}^{n \times n}$ and $Q \in \mathbb{R}^{n \times n}$.

Note that $f$ may be easily evaluated in $O(n^2)$ time by multiplying $y$ first by $W$, obtaining $z$, and then multiplying $z$ by $Z$, obtaining $Zz$, and finally pre-multiplying by $z^\top$ obtaining $z^\top Z z = y^\top W^\top Z W y$. Thus applying CP is relatively straight-forward, with the only potential difficulty being that the matrix $Z$, which is the local Hessian associated with the node that computes $z^\top Z z$, may not be easy to factorize. But using the $T/U$ variant of CP gets around this issue, and achieves a $O(n^2)$ computational cost. Moreover, it is easy to see how the required passes could be implemented by a fixed-depth arithmetic circuit (with gates of arbitrary fan-in) with $O(n^2)$ edge-cost since the critical operations required are just a few matrix-vector multiplications. The goal of the next theorem is to show that there can be no such circuit of edge cost $O(n^2)$ for computing the exact Hessian of $f$.

**Theorem 6.1.** *Any family of bounded depth arithmetic circuits with arbitrary fan-in gates which computes the diagonal of $f$ given inputs $W$ and $Z$ will have an edge count which is superlinear in $n^2$.*

The basic idea of the proof is to use the existence of such a circuit family to construct a family of circuits with bounded depth and edge count $O(n^2)$, that can multiply arbitrary $n \times n$ matrices (which will turn out to be the matrices $P$ and $Q$ that parameterize $f$), contradicting a theorem of Raz and Shpilka (2001) which shows that any such circuit fam-ily must have edge count which is superlinear $n^2$. The following lemma accomplishes this construction:

**Lemma 6.2.** *If an arithmetic circuit with arbitrary fan-in gates computes the diagonal of the Hessian of $f$ for arbitrary $P$, $Q$ and $Z$, then there is also a circuit of twice the depth $+ O(1)$, and three times the number of edges $+ O(n^2)$, which computes the product $PQ$ for arbitrary input matrices $P, Q \in \mathbb{R}^{n \times n}$.*

The results presented in this section rule out, or make extremely unlikely, the possible existence of algorithms which could perform a constant number of backwards and forwards "passes" through the computational graph of $f$ to find its exact Hessian.

## 7. Related work

The simplest way of computing the entries of the Hessian, including the diagonal, is by using an algorithm for Hessian-vector multiplication and running through the vectors $e_i$ for $i = 1...n$, recovering each column of $H$ in turn. Unfortunately this method is too expensive in most situations, and in the example function $f$ used in Section 6, would require $O(n^3)$ time.

The method of Chapelle and Erhan (2011) can be viewed as a special case of CP, where all the $M_i$'s except for the $M_i$ associated with the final nonlinearity are set to zero. Because of this, all of the results proved in this paper also apply to this approach, but with the Hessian replaced by the Gauss-Newton matrix.

Becker and Le Cun (1988) gave an approach for approximating the diagonal of the Hessian of a neural network training objective using a deterministic algorithm which does several passes through the computation tree. This method applies recursions similar to (4)-(6), except that all the "intermediate Hessians" at each layer are approximated by their diagonals, thus producing a biased estimate (unless the intermediate Hessians really are diagonal). We numerically compare CP to this approach in Section 8.

In Bishop (1992), a method for computing entries of the Hessian of a feedforward neural network was derived. This method, while being exact, and more efficient than the naive approach discussed at the start of this section, is not practical for large networks, since it requires a number of passes which will be at least as big as the total number of hidden and outputs units. CP by contrast requires only 1 pass to obtain a single unbiased rank-$B$ estimate, where $B$ is the number of training cases.

## 8. Experiments

### 8.1. Accuracy Evaluation

In this section we test the accuracy of CP on a small neural network as we vary the number of samples. The network consists of 3 hidden layers, each with 20 units. The input and output layers are of size 256 and 10 respectively giving a total of 6190 parameters. We tested both a network with random weights set by Gaussian noise with a variance of 0.01, and one trained to classify handwritten digits from the USPS dataset [2]. For the random vectors $v$, we tested both Gaussian and $\{-1, 1\}$-Bernoulli noise using the CP estimators based on using $S$ and $T/U$, and the simpler estimator discussed in Section 3 based on using $H/I$. For the sake of comparison, we also included the deterministic method of (Becker and Le Cun, 1988). The experiments were carried out by picking a subset of 1000 data points from the USPS dataset and keeping it fixed. Note that sample size refers to the number of random vectors generated *per data case*. This means that a sample size of 1 corresponds to an aggregation of 1000 rank-1 estimates.

Our results in 8.1 show that the accuracy of each estimator improves by roughly an order of magnitude for every order of magnitude increase in samples. It also shows that the $S$-based estimator along with binary noise is by far the most efficient and the simple $H/I$ based estimator is the least efficient by an order of magnitude.

### 8.2. Score-Matching Experiments

To test the effectiveness of CP in a more practical scenario, we focus on estimating the parameters of a Markov random field using the score matching technique. Score matching is a simple alternative to maximum likelihood that has been widely used to train energy-based models (Köster and Hyvärinen, 2007; Swersky et al., 2011). One of its drawbacks is that the learning objective requires the diagonal Hessian of the log-likelihood with respect to the data, which can render it unreasonably slow for deep and otherwise complicated models.

Our specific test involves learning the parameters of a covariance-restricted Boltzmann machine (cRBM; Ranzato et al., 2010). This can be seen as a two-layer network where the first layer uses the squared activation function followed by a second layer that uses the *softplus* activation function: $\log(1 + \exp(x))$. The details of applying score matching to this model can be found in Swersky et al. (2011).

In this experiment, we attempted to train a cRBM using stochastic gradient descent on minibatches of size 100. Our setup is identical to Ranzato et al. (2010). In particular, our cRBM contained 256 factors and hidden units. We trained the model on 11000 image patches of size $16 \times 16$ from the Berkeley dataset [3]. For our training procedure, we optimize the first layer for 100 epochs, then freeze those weights and train the second layer for another 25 epochs.

Score-matching requires the derivatives (w.r.t. the model parameters) of the sum of the diagonal entries of the Hessian (w.r.t. the data). We can thus use CP to estimate the score-matching gradient by applying automatic differentiation to the CP estimator itself (sampling and then fixing the random noise $V$), exploiting the facts that the linear sum over the diagonal respects expectation, and the derivative of the expectation over $V$ is the expectation of the derivative, and so this will indeed produce an unbiased estimate of the required gradient.

---

[2] http://cs.nyu.edu/~roweis/data/usps_all.mat

[3] http://www.cs.berkeley.edu/projects/vision/grouping/segbench

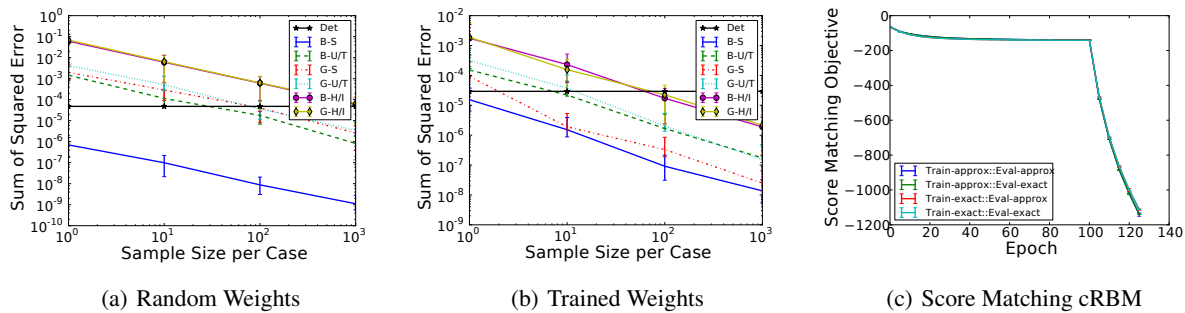

(a) Random Weights   (b) Trained Weights   (c) Score Matching cRBM

*Figure 1.* 1(a)-1(b): Accuracy of various estimators for the diagonal Hessian of a small neural network as the number of randomly drawn vectors per data case increases. B and G indicate the type of noise used (Binary or Gaussian), *S* and *U/T* are the complex and non-complex variants of CP, *H/I* is the simple approach discussed in Section 3, and Det is the approach of Becker and Le Cun (1988). 1(c): Score matching loss versus epoch when training using exact minibatch gradient and approximate minibatch gradient. In addition, when training with exact or approximate methods we also evaluate and plot the approximate/exact objective to ensure that they are not too different. Training the second layer begins after epoch 100.

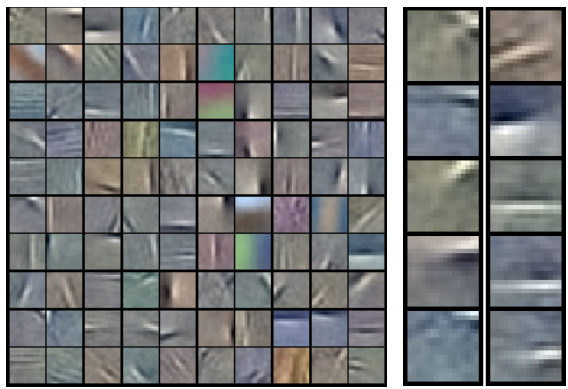

*Figure 2.* Covariance filters (left) and examples of second-layer pooling (right) from a cRBM learned with score matching on natural image patches using a stochastic objective.

A random subset of covariance filters from the trained model are shown in Figure 8.2. As expected the filters appear Gabor-like, with various spatial locations, frequencies, and orientations. The second layer also reproduces the desired effect of pooling similar filters from the layer below.

To demonstrate that learning can proceed with no loss in accuracy we trained two different versions of the model, one where we use the exact minibatch gradient, and one where we use approximate gradients via our estimator. We plot the training loss versus epoch, and our results in Figure 1(c) show that the noise incurred from our unbiased approximation does not affect accuracy during learning with minibatches. Unfortunately, it is difficult to train for many epochs in the second layer because evaluating the exact objective is prohibitively expensive in this model.

ACKNOWLEDGEMENTS

We thank Olivier Chapelle for his helpful discussions.